\documentclass[10pt,twocolumn,letterpaper]{article}

\usepackage{cvpr}
\usepackage{times}
\usepackage{epsfig}
\usepackage{graphicx}
\usepackage{amsmath}
\usepackage{amssymb}
\usepackage[utf8]{inputenc} 
\usepackage[T1]{fontenc}    
\usepackage{url}            
\usepackage{amsfonts}       
\usepackage{nicefrac}       
\usepackage{microtype}      
\usepackage{array,multirow,graphicx}
\usepackage{color}
\usepackage{booktabs,siunitx} 
\usepackage{makecell}
\usepackage{enumitem}
\usepackage{amssymb}
\usepackage{pifont}
\newcommand{\cmark}{\ding{51}}%
\usepackage{adjustbox} 
\usepackage{subcaption}
\usepackage[dvipsnames]{xcolor}

\usepackage[pagebackref=true,breaklinks=true,letterpaper=true,colorlinks,bookmarks=false]{hyperref}

\cvprfinalcopy 

\def\cvprPaperID{4354} 

\ifcvprfinal\fi
\begin{document}

\title{Sound and Visual Representation Learning with Multiple Pretraining Tasks}

\author{Arun Balajee Vasudevan$^{1}$, \hspace{.5cm}  Dengxin Dai$^{2}$, \hspace{.5cm}  Luc Van Gool$^{1,3}$  \\
ETH Zurich$^{1}$ \hspace{1.5cm} MPI for Informatics$^{2}$ \hspace{1.5cm}  KU Leuven $^{3}$  \\
{\tt\small \{arunv,vangool\}@vision.ee.ethz.ch, ddai@mpi-inf.mpg.de}
}

\maketitle

\thispagestyle{empty}

\begin{abstract}
Different self-supervised tasks (SSL) reveal different features from the data. The learned feature representations can exhibit different performance for each downstream task.
In this light, this work aims to combine Multiple SSL tasks (Multi-SSL) 
that generalizes well for all downstream tasks.
Specifically, for this study, we investigate binaural sounds and image data in isolation.
For binaural sounds, we propose three SSL tasks namely, spatial alignment, temporal synchronization of foreground objects and binaural sounds and temporal gap prediction. 
We investigate several approaches of Multi-SSL and give insights into the downstream task performance 
on video retrieval, spatial sound super resolution, and semantic prediction using OmniAudio dataset. 
Our experiments on binaural sound representations demonstrate that Multi-SSL via incremental learning (IL) of SSL tasks outperforms single SSL task models and fully supervised models in the downstream task performance. 
As a check of applicability on other modalities, we also formulate our Multi-SSL models for image representation learning and we use the recently proposed SSL tasks, MoCov2 and DenseCL. 
Here, Multi-SSL surpasses recent methods such as MoCov2, DenseCL and DetCo by $2.06\%$, $3.27\%$ and $1.19\%$ on VOC07 classification and +$2.83$, +$1.56$ and +$1.61$ $AP$ on COCO detection. Code will be made publicly available.

\end{abstract}

\section{Introduction}

\begin{figure}[t] \vspace{-5mm}
\centering
\includegraphics[width=1.0\linewidth, height=0.60\linewidth,trim={.18\textwidth} {.08\textwidth} {0.28\textwidth} {.12\textwidth},clip]{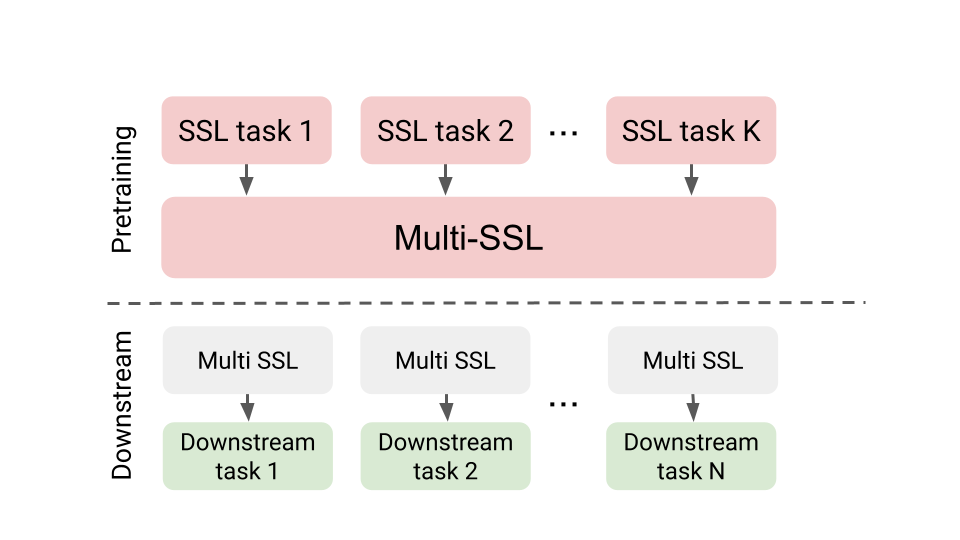} \\
\text{(a) Pipeline of our approach} \\
\includegraphics[trim=2 15 2 25,clip,width=1.0\linewidth]{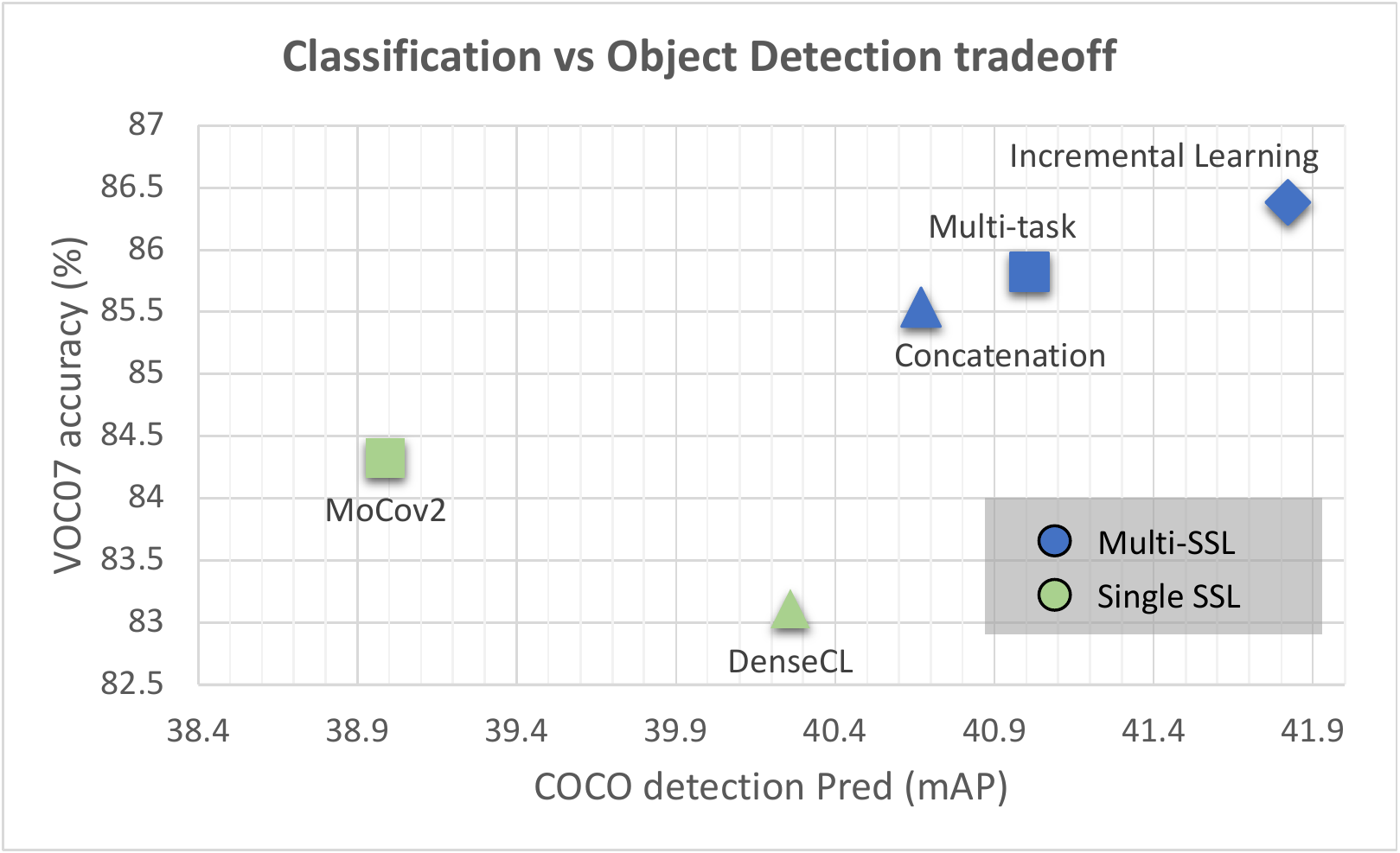} \\
\text{(b) Classification vs Object detection tradeoff} \\ \vspace{0mm}
\caption{Self Supervised Learning (SSL) tasks are designed for specific downstream tasks. 
Our work demonstrates how a single model learns to combine Multiple SSL tasks (i.e., Multi-SSL) that generalizes well for all the downstream tasks. Figure (b) shows comparison of single SSL tasks: MoCov2 \cite{he2020momentum} and DenseCL \cite{wang2021dense} with our Multi-SSL models.}
\label{fig:teaser}
\end{figure}

Self-supervised learning (SSL) is a popular paradigm to train deep networks on pretext tasks that readily extract supervision from the data. Typically, this allows models to learn data representations, which are then used for downstream tasks. 
The objectives for these SSL tasks are designed based on the corresponding downstream tasks. As a result,
pretrained deep networks on SSL tasks yield good performance on that downstream task or its related tasks when finetuned.
We note in the literature \cite{doersch2017multi,wang2021dense} that these pretrained models are not generic enough to give a satisfactory performance on a diverse pool of downstream tasks. For instance, some pretext tasks \cite{chen2020simple,he2020momentum} on images focus on learning global image feature representation while few others \cite{chaitanya2020contrastive,li2021dense,wang2021dense,roh2021spatially} focuses on local features. The former works well on downstream tasks like image retrieval or classification while the latter helps with dense prediction/labelling tasks. For instance, this is evident in Figure \ref{fig:teaser}(b) where MoCov2 \cite{he2020momentum} performs good for classification while DenseCL \cite{wang2021dense} comes out better for object detection task.

Be it visual, sound, or linguistic data, how well the data representations are learned determines the generalization of a model. When the models are generic, feature representations from them perform satisfactorily on several diverse downstream tasks.
In this light, our work tries to investigate how to train a self-supervised model using Multiple SSL tasks (Multi-SSL as in Figure \ref{fig:teaser}(a)) that generalizes well.

In the last few years, a number of self-supervised approaches are proposed in the language, sound and vision research community, from natural language text corpus \cite{devlin2018bert,peters2018deep,radford2018improving}, images \cite{li2021dense,jing2020self}, videos \cite{afouras2020self,han2020self,korbar2018cooperative}, and audios \cite{giri2020self,kuleshov2017audio}. In sound representation learning, a few prominent ones are audio-visual correspondence \cite{aytar2016soundnet,zhao2018sound}, audio context prediction \cite{tagliasacchi2019self}, and among others. Colorization \cite{larsson2017colorization}, image inpainting \cite{pathak2016context}, \textit{etc.,} are among the vision tasks for image representations. To assess the model, a standard set of downstream tasks for their corresponding areas is picked, and the model is retrained and tested on several datasets. 

With the aim of extracting well generalized sound and image representations, this paper explores ways of combining \textit{Multiple SSL} tasks as shown in Figure \ref{fig:teaser}(a). We call it Multi-SSL. 
While there have been a few works in this area on sound representation \cite{ravanelli2020multi,zaiem2021pretext,wu2021multi}, language \cite{wang2020multi} or visual representation learning \cite{doersch2017multi,ghiasi2021multi}, they only have only considered addressing this in the standard multi-tasking framework \cite{doersch2017multi,wang2021dense}. 
This work, however, introduces Multi-SSL which investigates different design options to combine multiple SSL tasks and provide insights into the downstream tasks. Through this comprehensive analysis, 
we highlight how Multi-SSL models improve over strong baselines in the evaluation of different downstream tasks. In the paper, we experiment with binaural sound and image data representations.

For binaural sound representation learning, we propose a set of SSL tasks. Firstly, spatial alignment task is proposed for learning spatial features in sounds. The task leverages the correspondence between binaural sounds and the rich spatial cues present in \ang{360} videos. Our second task is to learn temporal synchronization of moving objects in the scene and binaural sounds. We call this task as foreground alignment as it learns to align foreground objects and sounds. The third task of temporal gap prediction encourages the sound models to learn a sense of time gap between binaural sounds. For training the above tasks, we use the OmniAudio dataset \cite{vasudevan2020semantic} and evaluate the performance on three downstream tasks: a) video retrieval, b) auditory semantic prediction and c) spatial sound super resolution (S$^3$R).  

In addition, we examine the performance of the proposed Multi-SSL approach to visual representation learning. For SSL tasks, we consider the recently proposed contrastive learning paradigms on representation learning. Following MoCoV2 \cite{he2020momentum}, the first SSL task works with contrastive learning at the level of global image features. For the second task, we select dense contrastive learning \cite{wang2021dense} which focuses on the local features. We train the above SSL tasks on the ImageNet dataset \cite{deng2009imagenet} and then evaluate the performance on 
downstream tasks of image classification on Pascal VOC dataset  \cite{everingham2010pascal} and object detection and instance segmentation on MS COCO dataset \cite{lin2014microsoft}.

Furthermore, we propose different Multi-SSL methods such as Concatenation, Multi-task, ProgressiveNet, Incremental Learning (IL) and others, that are detailed in Section \ref{sec:multi_ssl}. Experiment results show that a) All the above Multi-SSL methods improve over single SSL tasks, b) and they also outperform supervised models, and finally c) IL approach performs the best among the Multi-SSL methods as in Figure \ref{fig:teaser}(b). We note that these observations are consistent for both sound and vision domains.

Here is a summary of our contributions. (1) We propose different approaches to self-supervised learning (SSL) for binaural sound representation learning; (2) We introduce several approaches of Multi-SSL that learns 
to combine multiple SSL tasks; (3) We also train and evaluate our Multi-SSL approach for image representation learning. 


\section{Related Works}

\noindent
\label{ref:relatedworks:SSL}
\textbf{Self-supervised learning.} 
Recently, SSL has become a key component to achieve good performance on downstream tasks predominantly with low-resource settings either in sounds 
\cite{giri2020self,kuleshov2017audio}, natural language processing \cite{lan2019albert,chen2020big} or computer vision \cite{misra2020self,jing2020self,li2021dense}. Let us focus more on sound and image representation learning in this work. In vision research, many
self-supervision tasks have been applied as a counter to ImageNet \cite{deng2009imagenet} pretraining.
Early self supervised pretext tasks typically include image colorization \cite{larsson2017colorization,zhang2016colorful}, orientation prediction \cite{gidaris2018unsupervised}, affine transform prediction \cite{zhang2019aet}, predicting contextual image patches \cite{doersch2015unsupervised}, reordering image patches \cite{carlucci2019domain}, counting visual primitives \cite{noroozi2017representation}. 
These pretext tasks typically predict some low-level image properties resulting in feature representations \textit{i.e.,} covariant to image transformations. 
Recently, contrastive learning gained considerable traction in SSL \cite{hjelm2018learning,bachman2019learning,oord2018representation,henaff2020data}, which drives the concept of maximizing the similarity of a representation across views while minimizing its similarity with distracting negative samples \cite{he2020momentum,chen2020simple,wu2018unsupervised}. Here, the positive pairs are usually created with multiple augmented views of the same image, while negative pairs are created from different images. However, there are few works \cite{chen2021exploring,grill2020bootstrap} that use just positive samples. Our work explores several of these contrastive learning based SSL such as \cite{he2020momentum,wang2021dense,henaff2021efficient,roh2021spatially} and its variants for learning initial image representations. We pick MoCov2 \cite{he2020momentum} and DenseCL \cite{wang2021dense} in our work.

\smallskip
\noindent
\textbf{Audio-visual learning.} Audio-visual data offers a variety of resources for knowledge transfer between different modalities \cite{aytar2016soundnet,castrejon2016learning,afouras2020self}. Many works \cite{vasudevan2020semantic,arandjelovic2018objects} leverage the natural synchronization between vision and sound to learn representation of sounds and images without ground truth labels.
This has been successfully used in various tasks such as visually guided sound source separation \cite{gao2018learning} and sound localization \cite{arandjelovic2018objects}, audio to visual generation \cite{zhou2019talking}, visual to audio generation\cite{zhou2018visual,owens2016visually,ephrat2017vid2speech}, sound inpainting \cite{zhou2019vision,marafioti2019context} and sound classification \cite{aytar2016soundnet}. Prior works are often implemented either by predicting audio-visual correspondences at the video level \cite{arandjelovic2017look,morgado2021audio}, frame level and object level alignment \cite{afouras2020self}. Han \textit{et al} \cite{han2020self} uses optical flow patterns in video to learn video representations. We inherit these flow patterns to find foreground objects and sound correspondence to learn sound representations.
Morgado \textit{et al} \cite{morgado2020learning} learn representations by performing audio-visual spatial alignment of \ang{360} video and spatial audio. 
In our spatial alignment SSL, we differ with them in learning binaural sounds representation. \cite{tagliasacchi2019self,wei2018learning} propose prediction of sense of time difference as SSL task, given two video/audio frames, to learn video/audio representations. We follow quite similar to the work of \cite{tagliasacchi2019self}.

\smallskip
\noindent
\textbf{Multi-task self-supervised learning.} While it has been shown extensively in  supervised learning settings \cite{mtl:survey,hoyer2021three}, the literature on multi-tasking in SSL remains less explored. There are extensive studies on pretext tasks as we see in Section \ref{ref:relatedworks:SSL} for image or sound representation learning. In addition, a few SSL works \cite{pascual2019learning,ravanelli2020multi,lee2019multi} in computer vision and speech address combining multiple pretext SSL tasks in a  multi-task setting. Successful pretext tasks such as Jigsaw \cite{doersch2015unsupervised}, colourisation and rotation \cite{gidaris2018unsupervised} have been combined successfully to improve downstream performance \cite{kim2018learning}.
Wang \textit{et al} \cite{wang2021dense} employs contrastive learning paradigm at image level and dense level features in multi-tasking settings. In our work, we explore more ways to combine these paradigms and explore how incremental learning helps in this context.

\section{Self-Supervised Tasks}
\label{sec:SSL}

We use different set of pretext tasks for learning binaural sounds and image representations. We pick diverse set of SSL tasks aiming to extract diverse feature representations.
These are more likely to span the space of features needed to understand general data content. Initially, let us see SSL pretext tasks for binaural sounds in Section \ref{sec:SSL_binaural}, and then for visual representations in Section \ref{sec:SSL_images}.

\subsection{Binaural sounds}
\label{sec:SSL_binaural}

Given an audio-visual dataset with $N$ raw video segments, e.g. $D = \{(a_{1},v_{1}), (a_{2},v_{2}), ... , (a_{N},v_{N}) \}$, the objective for SSL task is to obtain a function $f(.)$ that is effectively used to generate sound representations for various downstream tasks.
In our work, we formulate sound clips as spectrograms, which are effectively processed by convolutional neural networks(CNNs) as demonstrated by \cite{gao20192,arandjelovic2018objects}.
Let us see the SSL pretext tasks.


\smallskip
\noindent
\textbf{Spatial alignment} (denoted as $\mathcal{A}$). 
This pretext task learns to align \ang{360} videos spatially with their corresponding binaural sounds.
A straightforward way to implement audio-visual spatial alignment is to rotate the video randomly with $R$ angle rotation with respect to sounds to create an artificial misalignment between them. And, later we learn to predict this rotation angle between the video and the sounds.
In learning the spatial alignments of visual and sound contents, the network is encouraged to understand the scene composition (\textit{i.e.,} where the different sources of sound are located), which results in better representations for downstream tasks. Closest work to this pretext task is \cite{morgado2020learning} which employs contrastive learning setup to learn the spatial alignment between \ang{360} videos and spatial sounds. In our work, we leverage the pairs of binaural sounds and \ang{360} videos and frame the problem as an angle prediction between them. 
We divide \ang{360} into $8$ equal bins, each representing different orientations.
We train the SSL model to predict the angular difference \textit{i.e.,} rotation angle $R$. We employ cross entropy loss between the predicted and actual angular difference as $CE(
    h(v_{iR},a_{i}), R)$,
where $h(v_{iR},a_{t})$ is a prediction head followed by a softmax layer to predict the angle $\hat{R}$. $v_{iR}$ and $a_{i}$ refer to video features of rotated video segment $v_{i}$ and sound features respectively. $R$ is the groundtruth angular difference 
due to the rotation 
on \ang{360} video segment $v_{i}$. $CE$ denotes the cross entropy loss.

\smallskip
\noindent
\textbf{Foreground alignment} (denoted as $\mathcal{B}$). Using unlabeled videos, our work aims to harness the natural synchronization of vision and sound to learn binaural sound representations. 
Audio-visual temporal synchronization (AVTS) \cite{korbar2018cooperative,owens2018audio} 
distinguishes between a pair of audio and video clips belongs to the same timestamps (aligned) or from separate timestamps (misaligned) of the same video. 
We leverage this alignment to train our model in a contrastive learning setup.

The motion of objects and their sounds are closely related. In this light, we apply two approaches based on the fair assumption that the recorded sounds from the scene come from moving foreground objects alone. 
First, we extract the spatial masks of foreground objects outlined in \cite{vasudevan2020semantic}.
Features of the masked foreground objects are learned to align with sounds similar to AVTS case. 
The results of this first part are in the supplementary material. 
Secondly, we use motion flow features. Based on our assumption, all the sound making objects in the scene are in motion. Indeed, this promotes self-supervision by aligning motion flow features with sound features. Our experiments with second approach are discussed in Section \ref{sec:experiments}. 
Coming to the loss, we define it as: $-log\dfrac{exp(v_{p} \cdot a_{i}/\tau)}{exp(v_{p} \cdot a_{i}/\tau) + \sum_{v_{n} \in P_{i}}exp(v_{n}\cdot a_{i}/\tau)}$

where $v_{p}$ and $v_{n}$ are video feature vectors from aligned (positive) and misaligned (negative) video segments respectively, with respect to sound segment feature $a_{i}$, and $\tau$ is a temperature hyperparameter. 
$P_{i}$ represents the set of misaligned video features for $a_{i}$.




\smallskip
\noindent
\textbf{Temporal gap prediction} (denoted as $\mathcal{C}$).
This pretext task consists of estimating the time difference between any two sound segments that are randomly sliced from a single longer sound clip.
For the task, we frame a model that takes $2$ sampled sound segments as inputs and learn to estimate the distance in time between them quite similar to \cite{tagliasacchi2019self}.
Specifically, let us assume the length of sliced sound clip be $T$ and original clip be $T_{max}$.
We extract two sound slices $a_{i}$ and $a_{j}$ such that $\Delta = |t_{i} - t_{j}|$. Here, $t_{i}$ and $t_{j}$ are timestamps of $a_{i}$ and $a_{j}$ and $\Delta$ is sampled from a uniform distribution, $U(0,T_{max} -T)$. This temporal gap between sound slices is normalized as $\delta = \Delta/(T_{max} -T) \in [0, 1]$.
It is important to note that there is no temporal order between the two slices. We concatenate the sound representations $[a_{i},a_{j}]$ into a single vector and we feed this vector into a fully connected feed forward network with a single hidden layer of size $64$ that produces the scalar output $\hat{\delta} $. We train the model end-to-end so as
to minimize a huber loss $L_{gap}(\delta,\hat{\delta}) $  between the ground-truth and the predicted temporal gap. 

\begin{figure*}[!tb]
\begin{tabular}{ccc}
  \includegraphics[trim=8 5 8 0,clip,width=0.34\textwidth,height=0.18\textwidth]{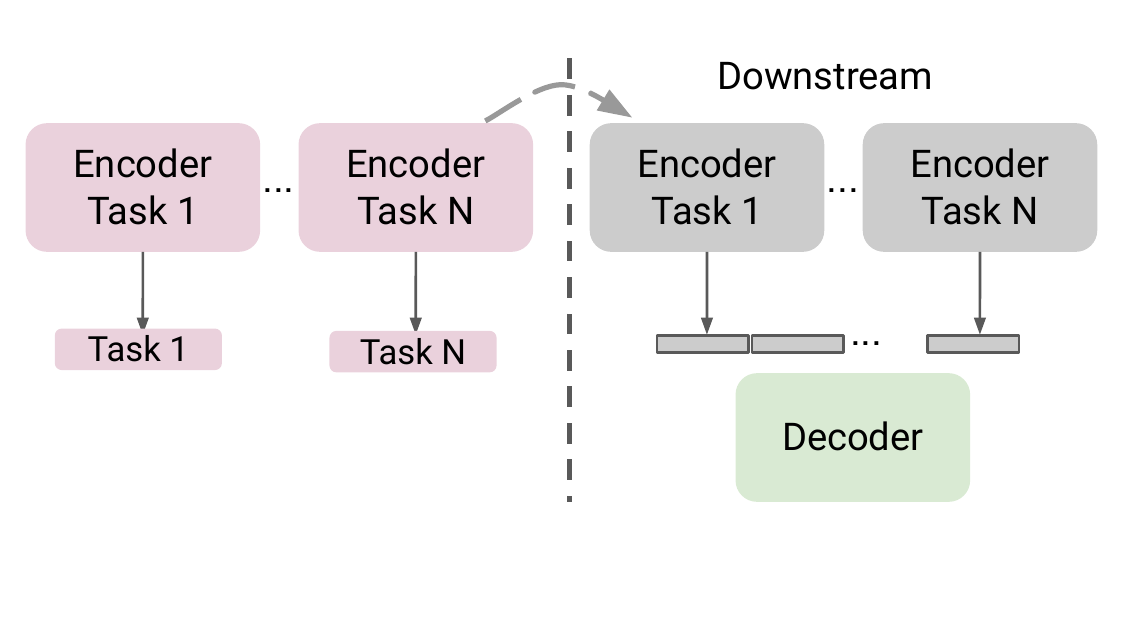} & 
  \includegraphics[trim=4 0 10 13,clip,width=0.290\textwidth,height=0.165\textwidth]{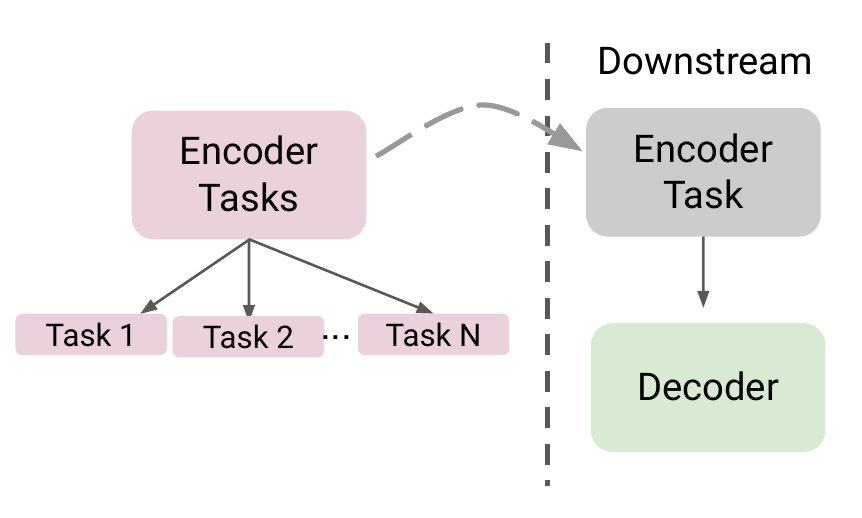} & \hspace{1mm}
  \includegraphics[trim=4 0 8 0,clip,width=0.32\textwidth,height=0.18\textwidth]{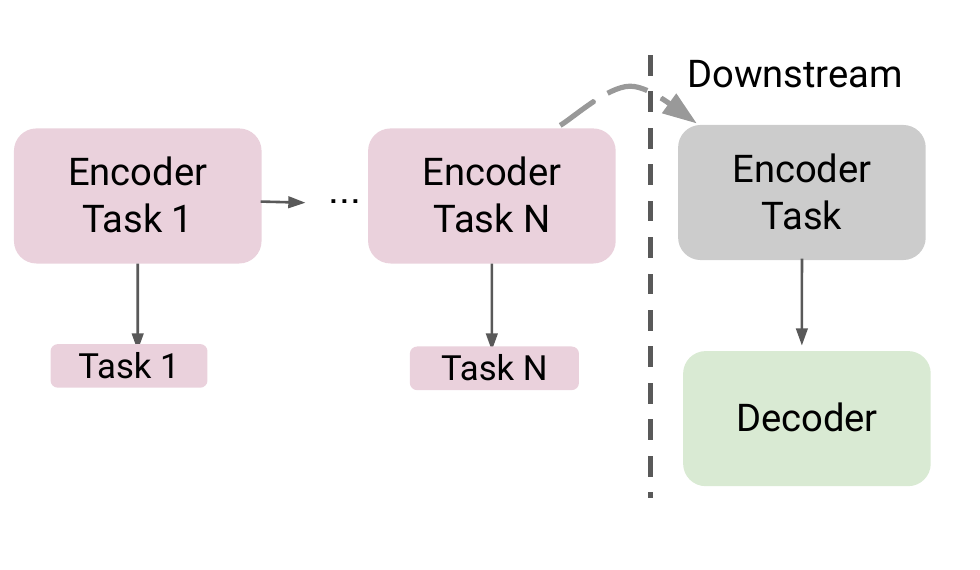}   
  \\ 
  \text{(a) Concatenation}  & 
  \text{(b)  Multi-tasking} & \hspace{-5mm}
  \text{(c) Incremental learning} 
  \\ 
  \end{tabular} \vspace{-2mm}
  \caption{Different ways of combining multiple self-supervised approaches. Left side of each subfigure indicates the Multi-SSL pretraining and right side depicts the downstream task training and evaluation. \textcolor{Gray}{Gray} blocks denote frozen part.} 
  \label{fig:multiple_ssl}
\end{figure*}

\noindent
\subsection{Images}
\label{sec:SSL_images}

A number of recent research on self supervised learning \cite{oord2018representation,wu2018unsupervised,bachman2019learning}
has demonstrated the benefits of using a discriminative contrastive loss on data samples.
Contrastive learning can drive a variety of pretext tasks and we choose a few of them, each carrying out different mechanisms.

\smallskip
\noindent
\textbf{MoCo.} Momentum Contrast (MoCo) applies contrastive loss to features at the image level. MoCo \cite{he2020momentum,chen2020improved} shows that unsupervised learning can be superior to its ImageNet-supervised counterpart in image classification and detection tasks.
MoCo uses two encoders, an encoder and a momentum encoder and the encoded representations are called queries and keys, respectively. 
MoCo trains a visual representation encoder by matching an encoded query q to a dictionary of encoded keys using a contrastive loss. 
We consider a query and a key as a positive pair if they originate from the same image and are two random views under random data augmentation, and otherwise as a negative sample pair. 
MoCo is designed for learning global representations, based on which the model is fine-tuned later for image classifications. 
Recent researches show that local features can also be extracted and compared using contrastive learning.


\begin{table}[!tb]
  \centering 
\begin{adjustbox}{max width=\textwidth,max totalheight=\textheight}
\resizebox{1.03\linewidth}{!}{
  \begin{tabular}{cccccccc}
\toprule
   Name & Train & Eval & Task & \# Size  \\ \midrule
   OmniAudio \cite{vasudevan2020semantic}& \cmark & & SSL & 64K  \\
   OmniAudio & &\cmark & SP, VR, S$^3$R & 64K \\ \midrule
   ImageNet \cite{deng2009imagenet}& \cmark & & SSL & 1.28M  \\
   Pascal VOC \cite{everingham2010pascal}& &\cmark & Image Classification & 16K \\
   COCO \cite{lin2014microsoft} & &\cmark & Detection, Inst Segm & 118K \\
\bottomrule
\end{tabular} 
}
\end{adjustbox} \vspace{-2mm}
\caption{Datasets used for Multi-SSL pretraining and for sound and visual downstream evaluation.}\label{tab:dataset_list}%
\end{table}

\smallskip
\noindent
\textbf{Dense contrastive learning.}
DenseCL \cite{wang2021dense} proposes a self-supervised learning framework to handle dense prediction/labelling tasks. DenseCL is primarily viewed as a dense pairwise contrastive learning as opposed to the global image representation learning. To begin with, a dense projection head is defined that takes the backbone features as inputs and then produces dense feature vectors. By producing a dense output format, we maintain spatial information unlike the existing global projection head that outputs a single, global feature vector for each image. Further, we determine the positive sample for each local feature vector by extracting the correspondence across views of the same image. We then construct an SSL loss function by extending the conventional InfoNCE loss \cite{henaff2020data} to a dense paradigm i.e., dense contrastive loss. We then perform contrastive learning densely using a fully convolutional network (FCN) \cite{long2015fully}, and the pretrained network is used to target dense prediction tasks.

\section{Multi-SSL pretraining}
\label{sec:multi_ssl}

Several existing self-supervised learning (SSL) approaches choose a self-supervision objective based on the downstream task.
The aim of our study is to determine whether we can combine Multiple SSL (Multi-SSL) tasks to simultaneously train a single encoder network. Furthermore, this encoder representations yield better downstream performance. Combining these tasks fairly in a multiple task learning objective is challenging 
and we discuss how we overcome this problem in subsequent sections. We call this approach as Multi-SSL. In our experiments, we investigate whether multiple SSL tasks extract general feature representations more effective than single task ones. 
Additionally, we examine which combination of SSL tasks are more beneficial and give a notable boost. In this section, we will discuss three prominent ways of combining multiple tasks we investigated and briefly summarize few other methods. Whenever possible, we follow the procedures established in the previous works, although in many cases modifications are necessary for our multiple task pretraining. 

Let us assume we have $K$ SSL tasks and $N$ downstream tasks as shown in Figure \ref{fig:teaser}. For single SSL task pretraining, we have different encoders $Enc_{k}$ trained for each task to yield feature representations $\{f_{1}, f_{2},..., f_{K}\}$. Let us denote decoder for $n_{th}$ downstream task $D_{n}$ as $d_{n}$. 

\subsection{Concatenation}
Firstly, let us consider a naive way of combining multiple tasks under Multi-SSL pretraining. Here, our approach is to concatenate the encoder features from $K$ encoders. We train specific encoder networks $Enc_{k}$ for specific SSL tasks separately. Each SSL task is designed to suit a particular downstream task $D_{n}$ as we see in Section \ref{sec:SSL}. Different features of encoder can complement each other. Here, we investigate the approach of concatenation of all SSL encoder features $f_{k}$ to combine multiple SSL tasks. Further, these concatenated features $	[f_{1}, f_{2},..., f_{K}]$ are frozen and passed to downstream task specific decoders $d_{n}$, where we just train the decoders as shown in Figure \ref{fig:multiple_ssl}(a). 
The downstream task decoder $d_{n}$ chooses the appropriate input features from concatenated features based on its task.
However, this approach has its own limitation. In proportion to the number of SSL tasks added to \textit{Concatenation}, the number of encoder trunks and concatenated features increase proportionally. 
In the next subsections, we can see how a single encoder trunk is learned for Multi-SSL approach.

\begin{table*}[!tb]
  \centering 
\caption{ Single-SSL in (a) and Multi-SSL methods in (b). All methods are trained and evaluated on OminAudio dataset. }\label{tab:sound:multiple_SSL}%
\begin{subtable}[h]{0.3\textwidth}
\resizebox{0.95\linewidth}{!}{
\begin{tabular}{ccccccccccc}
\toprule
   \multicolumn{1}{c}{SSL} &  \multicolumn{3}{c}{Downstream tasks} \\
 &  SP$\uparrow$ & S$^3$R$\downarrow$ & VR$\uparrow$ \\  \midrule
 Sup & 26.82 & 0.2085  & - \\
 \midrule
 $\mathcal{A}$  & 15.32 & \textbf{0.2105} & 9.13\\
 $\mathcal{B}$    & \textbf{24.33} &  0.2501 & \textbf{27.35}\\ 
 $\mathcal{C}$& 16.85 & 0.2931  & 20.44 \\ 
\bottomrule
\end{tabular}
}
\caption{Sound representations from 3 SSL tasks are evaluated on 3 downstream tasks. $\mathcal{A}$:Spatial alignment, $\mathcal{B}$: Foreground alignment, $\mathcal{C}$: Temporal gap prediction.}
\end{subtable}
\hspace{2mm}
\begin{subtable}[h]{0.65\textwidth}
\resizebox{1.03\linewidth}{!}{
\begin{tabular}{ccccccccccccccc}
\toprule
  \multicolumn{1} {c} {Multi-SSL} & \multicolumn{2}{c}{Semantic prediction$\uparrow$} & \multicolumn{2}{c}{Video Retrieval $\uparrow$}& \multicolumn{2}{c}{S$^3$R $\downarrow$}\\
 Methods   & $\mathcal{B}$+$\mathcal{C}$ & $\mathcal{B}$+$\mathcal{C}$+$\mathcal{A}$ & $\mathcal{B}$+$\mathcal{C}$ & $\mathcal{B}$+$\mathcal{C}$+$\mathcal{A}$ & $\mathcal{B}$+$\mathcal{C}$ & $\mathcal{B}$+$\mathcal{C}$+$\mathcal{A}$  \\  
 \midrule
 Euclidean dist   & 25.38 & 25.82 & 27.64 & 27.45& 0.2607 & 0.2188\\
 Contrastive dist  & 25.59 & 27.39 & 27.96 & 27.95& 0.2589 & 0.2145\\
 ProgressNet \cite{rusu2016progressive}   & 30.38 & 32.45 & 29.06 & 29.68 & 0.2397 & 0.2035\\
 Concatenate  & 26.37 & 30.14 & 28.31 & 28.09& 0.2495 & 0.2101\\
 Multi-task   &  27.28 & 31.21 & 28.94 & 29.52& 0.2411 & 0.2066\\
 IL    & \textbf{32.76} & \textbf{34.05} & \textbf{29.72} & \textbf{30.32} & \textbf{0.2378} & \textbf{0.1988} \\
\bottomrule
\end{tabular} 
}
\caption{ Different methods of Multi-SSL approaches are evaluated on two-task $\mathcal{B}$+$\mathcal{C}$ and three-task $\mathcal{B}$+$\mathcal{C}$+$\mathcal{A}$ combination.  }
\end{subtable}
\end{table*}

\subsection{Multi-task}
Multi-task setting occurs when multiple tasks are combined with the goal of improving all tasks simultaneously through the sharing of common knowledge.
Here, in our work, multiple SSL tasks (K tasks) are trained simultaneously to learn shared representations. As discussed in Section \ref{sec:SSL}, we have three SSL tasks for learning sound representations and two for learning image representations.
Inspired from the work of \cite{doersch2017multi}, we employ a base feature encoder trunk 
that are shared among all SSL tasks. The encoder features are further passed as input to the task-specific output heads as in Figure \ref{fig:multiple_ssl}(b). These output heads are subjected to different SSL losses as discussed in Section \ref{sec:SSL}.
Here, we assign each SSL approach a separate task and train them jointly in a multi-task setting. To train the model, we use the weighted sum of loss from all $K$ SSL tasks. 
Later, we use the encoder trunk of the trained model under multi-tasking for the downstream task training.


\subsection{Incremental learning (IL)}
Continual learning is another popular paradigm to learn new tasks one after the other, which we employ as another Multi-SSL approach for SSL tasks. Learning without forgetting \cite{li2017learning} focuses on learning new tasks while preserving the performance of old tasks. Inspired from them, we learn multiple SSL tasks 
sequentially in an incremental manner. This means that we keep the same base encoder trunk model for all $K$ tasks and attach task-specific output heads for each SSL task as shown in Figure \ref{fig:multiple_ssl}(c). We learn the first SSL method using its task-specific layers and the responses are saved. Keeping the pretrained base trunk from the first task, we add task-specific layers of the second task and train it. During this training of the second task, we also retrain the first task's specific layers with their old responses. This way, we train the base trunk along with task specific layers of first and second task and backpropagate the combined loss from both tasks. In the same manner, we continue adding more SSL tasks to Multi-SSL IL to learn incrementally. 
Upon completion of all the tasks, we use the base encoder trunk of the trained model for the downstream tasks as shown in Figure \ref{fig:multiple_ssl}(c). 

We use the same dataset for all $K$ SSL tasks in Multi-SSL, as shown in Table \ref{tab:dataset_list}. Details are in Section \ref{sec:experiments}. 
Hence, we store the output responses from task specific layers for all the tasks once the task is completed. 
Whenever we learn a new task with new task-specific layers, we use 2 kinds of losses.
For the loss of the current task, groundtruth output from the current task is used anad secondly, for the loss of all the previous tasks, we use their stored output responses as groundtruth. 

\subsection{Other methods}

As part of Multi-SSL, we also attempt other approaches. For \textit{Euclidean dist} and \textit{Contrastive dist} of Table \ref{tab:sound:multiple_SSL},  we learn separate models for each SSL task. Each data point in the dataset is passed through the learned model and feature representations are extracted for each SSL task and are stored. Then, we learn a new base encoder trunk model which is learnt to output feature representation using their corresponding losses.
We apply L2 loss between base trunk features and stored latent representations of $K$ SSL tasks to train the base model for Euclidean dist. In the case of \textit{Contrastive dist:}, contrastive loss is applied which pulls together base trunk features and stored SSL representations of positive pairs while pushing apart latent representations of misaligned data points.
Further, we investigate ProgressiveNet \cite{rusu2016progressive} as in Table \ref{tab:sound:multiple_SSL} which is another continual learning approach. For single SSL tasking, we adopt baseline2 approach in \cite{rusu2016progressive} and later we follow the same work to add more SSL tasks.

\section{Experiments}
\label{sec:experiments}
In this section, we elaborate the experiments on transferability of single SSL and Multi-SSL models to different downstream tasks. The details of sound and image downstream tasks are in the Section \ref{sec:soundexperiments} and Section \ref{sec:imageexperiments} respectively and we discuss about the results of single SSL in Section \ref{sec:sound:single_SSL} and Multi-SSL in Section \ref{sec:sound:multiple_SSL}. In addition, we conduct extensive ablation studies on combination of SSL tasks in Multi-SSL in Section \ref{sec:sound:ablation}.

\subsection{Sound Downstream tasks}
\label{sec:soundexperiments}
We apply three SSL tasks to extract initial sound representations, as discussed in Section \ref{sec:SSL_binaural}. We use OmniAudio dataset \cite{vasudevan2020semantic} for all SSL tasks. Later, for the evaluation of SSL and Multi-SSL models, we consider three downstream tasks as described below, which tests the diversity and generalizability of sound representations.

\noindent
\textbf{Settings.} We use OmniAudio dataset \cite{vasudevan2020semantic} for training and testing of downstream tasks. For all the experiments, we collect training and testing samples of 2-second video segments and a pair of binaural sound channels. We preprocess sound samples following techniques from ~\cite{gao20192,vasudevan2020semantic}. 
More details are added in the supplementary material.

\noindent
\textbf{Video retrieval} (VR) is a common downstream task that aims 
to retrieve relevant videos based on a given sound clip. Following standard practices, we extract the sound representations from pretrained models on OmniAudio dataset \cite{vasudevan2020semantic}, and measure the top-1 accuracies of retrieving the video segment, obtained for a single sound segment. 

\noindent
\textbf{Semantic prediction} (SP) \cite{vasudevan2020semantic} is a downstream task for binaural sounds which deals with the prediction of semantics of sound-making objects as pixel-level labelling task given the binaural sounds. We use the base encoder trunk model of sound network pretrained on SSL approaches and attach a decoder and train the whole model to predict the semantic segmentation masks of $5$ classes- bus, car, tram, motorcycle and trucks. 

\noindent
\textbf{Spatial sound super resolution} (S$^3$R) \cite{vasudevan2020semantic} is a downstream task which aims to increase the directional resolution of sounds. This can be another testbed for binaural sounds. Here again, we attach a specific decoder to the encoder trunk pretrained on SSL approaches. Later, the model is trained and evaluated for S$^3$R task.

\begin{table*}[!tb]
  \centering 
\caption{Ablation studies on different combination of SSL tasks are provided. SSL tasks are $\mathcal{A}$: Spatial alignment, $\mathcal{B}$: Foreground alignment, $\mathcal{C}$: Temporal gap prediction. Different methods of Multi-SSL approaches like Concatenation, Multi-task and Incremental Learning and for evaluation downstream task of semantic prediction, video retrieval and S$^3$R is considered. Arrows indicate whether higher or lower is better.    }\label{tab:sound:ablation:multissl}%
\begin{tabular}{ccccccccccccccc}
\toprule
  \multicolumn{3} {c} {Self-supervised tasks} & \multicolumn{3}{c}{Semantic prediction $\uparrow$} & \multicolumn{3}{c}{Video Retrieval $\uparrow$}& \multicolumn{3}{c}{S$^3$R $\downarrow$}\\
  $\mathcal{A}$  & $\mathcal{B}$ & $\mathcal{C}$  & Concat & Multi-task & IL  & Concat & Multi-task & IL   & Concat & Multi-task & IL  \\  
 \midrule
 \cmark & \cmark &    &  29.55 & 29.23 & 31.89 & 24.47& 25.10&  25.65 & 0.2165 & 0.2073 & 0.2029\\
 \cmark & & \cmark  & 23.57 & 24.73 & 25.02 & 19.03 & 20.22 &  21.00 & 0.2347& 0.2308 & 0.2251\\
  & \cmark & \cmark  & 26.37 & 27.28 & 32.76 & \textbf{28.31} & 28.94 & 29.72 & 0.2495 & 0.2411 & 0.2378\\
  \cmark & \cmark & \cmark  & \textbf{30.14} & \textbf{31.21} & \textbf{34.05} & 28.09 & \textbf{29.52} &  \textbf{30.32}  & \textbf{0.2101} & \textbf{0.2066} &  \textbf{0.1988}\\
\bottomrule
\end{tabular} 
\end{table*}


\begin{figure*}[!tb]
\begin{tabular}{ccc}
  \includegraphics[trim=15 2 8 25,clip,width=0.32\textwidth,height=0.19\textwidth]{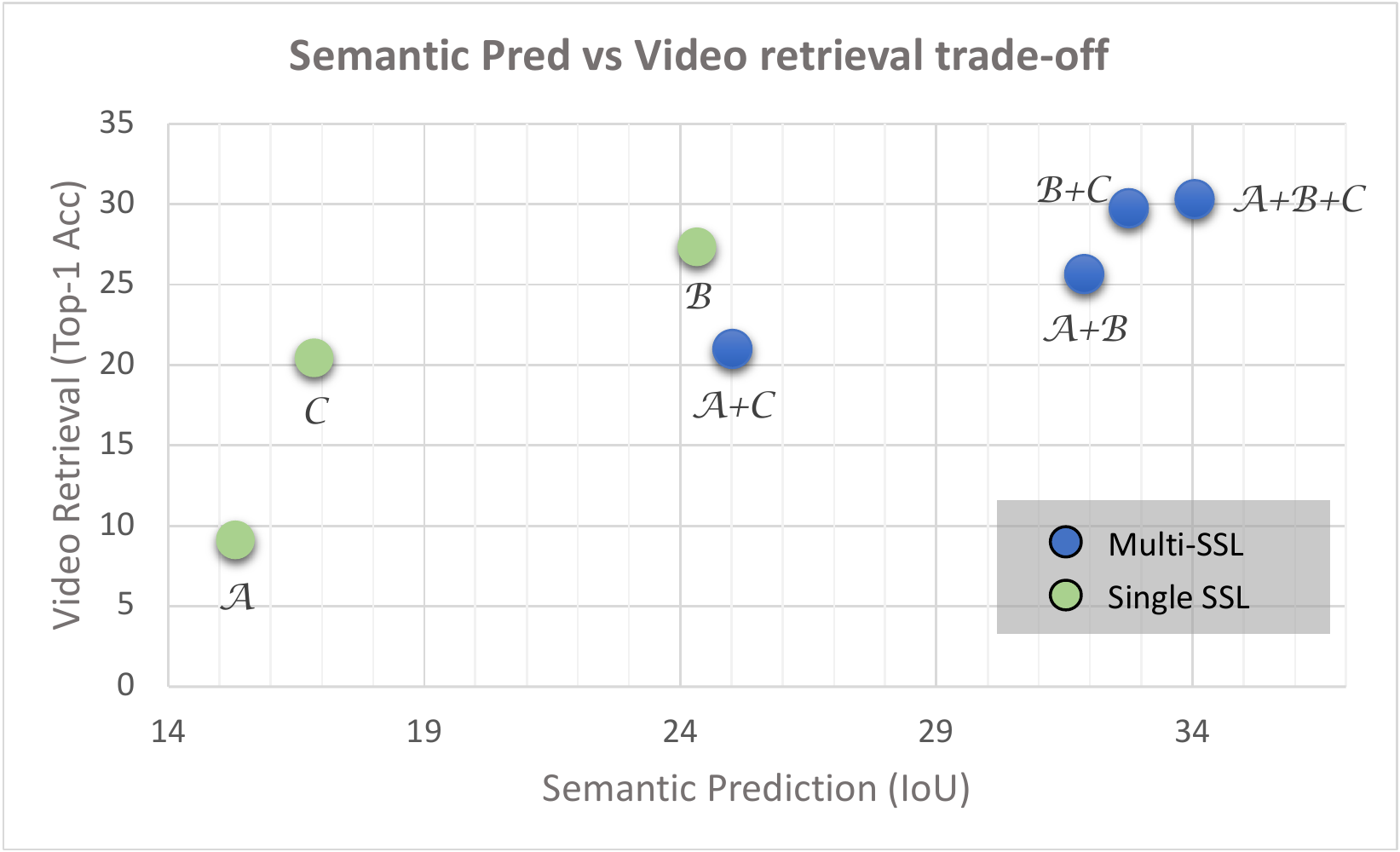} & \hspace{-2mm} 
  \includegraphics[trim=20 2 20 25,clip,width=0.32\textwidth,height=0.19\textwidth]{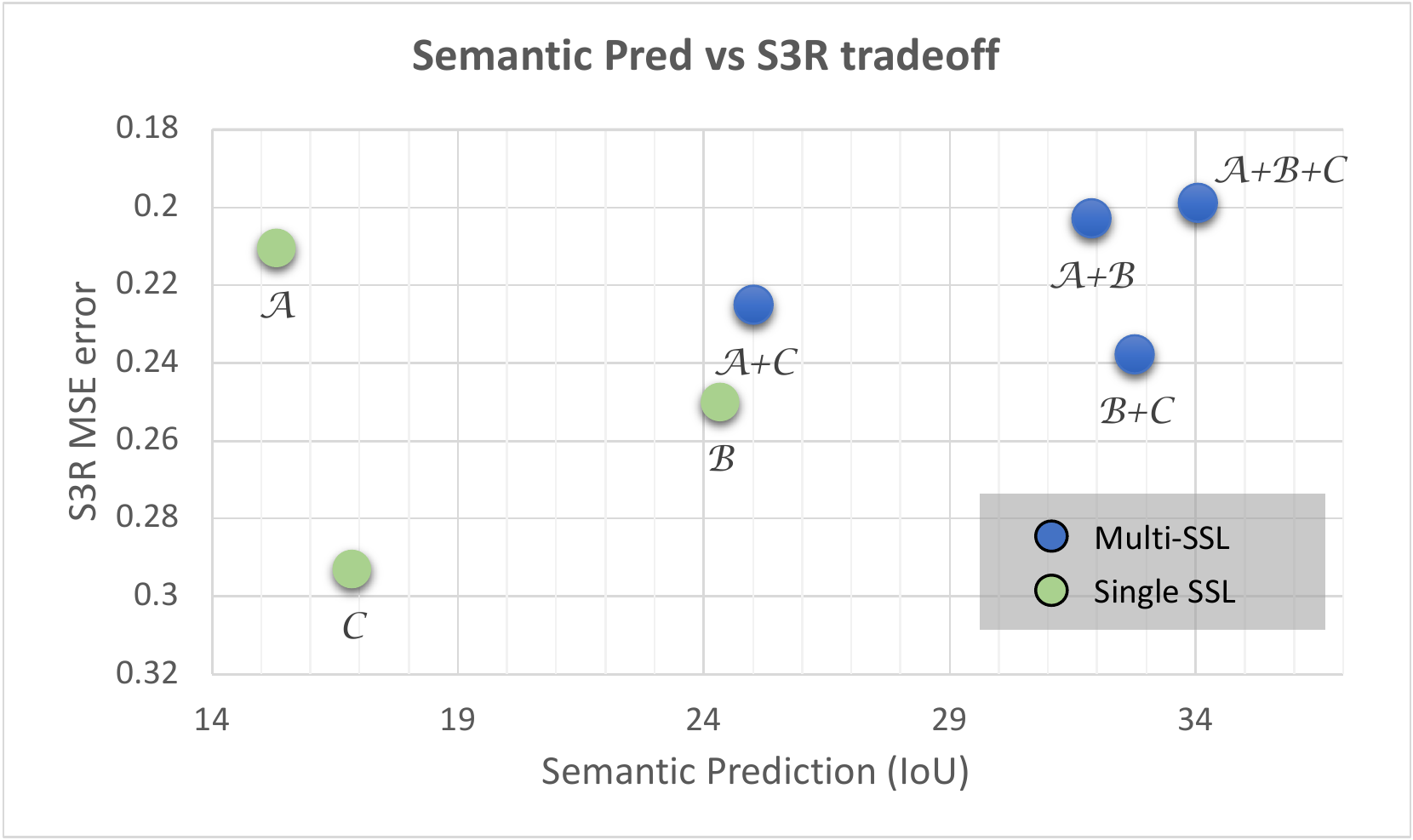} & \hspace{-2mm}
  \includegraphics[trim=2 2 3 25,clip,width=0.32\textwidth,height=0.19\textwidth]{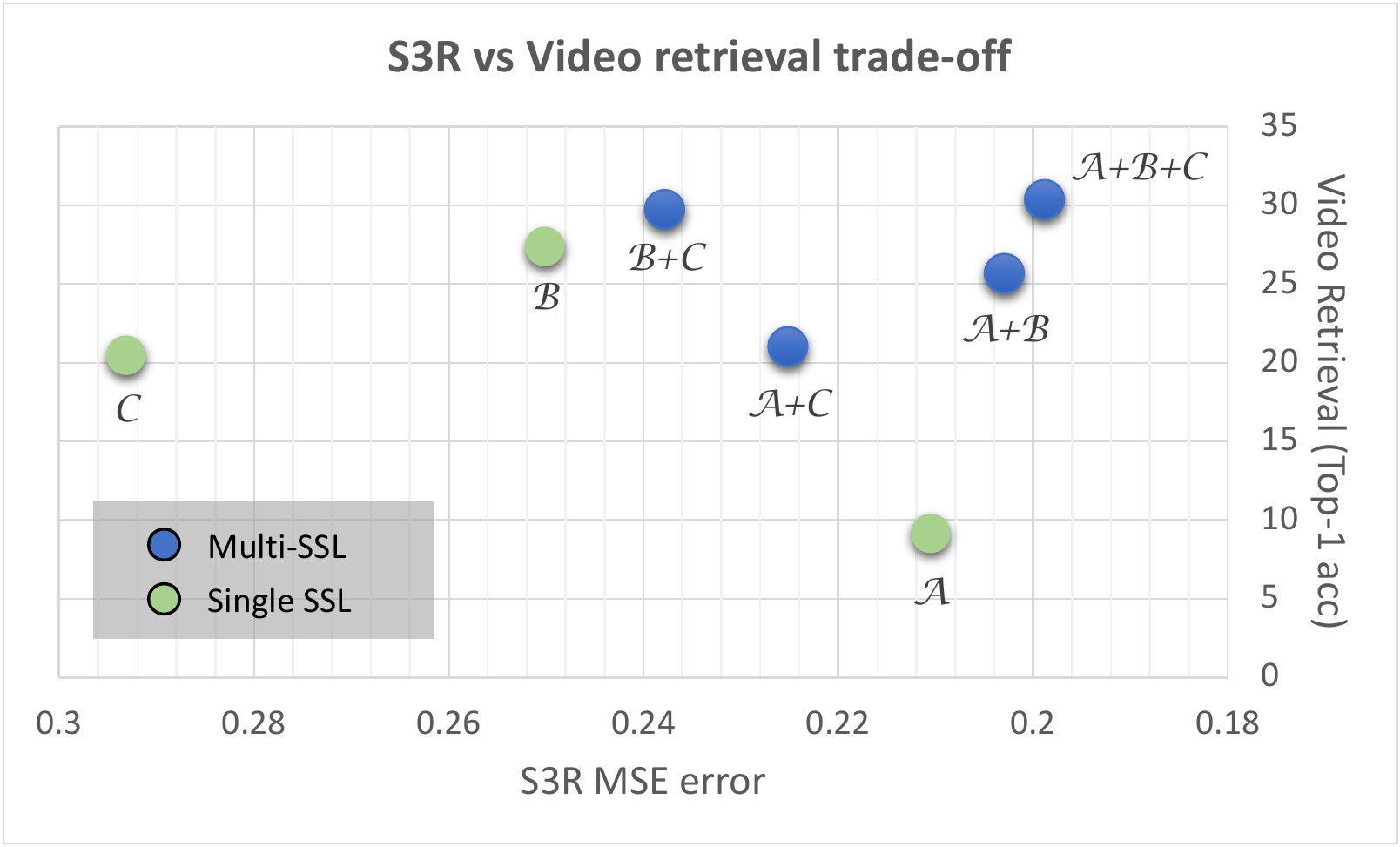}   
  \\ 
  \text{(a) Semantic Pred vs VR}  & \hspace{-2mm} \text{(b)  Semantic Pred vs S$^3$R} & \hspace{-5mm} \text{(c) S$^3$R vs VR}  \\ 
  \end{tabular} \vspace{-2mm}
  \caption{This presents the trade-off between semantic prediction, video retrieval accuracy and S$^3$R performance of our single SSL tasks (\textcolor{LimeGreen}{green}) and Multi-SSL approaches (\textcolor{MidnightBlue}{blue}). All the methods are trained and evaluated on OmniAudio dataset.}
  \label{fig:tradeoff_seman_VR}
\end{figure*}


\subsection{Visual Downstream tasks}
\label{sec:imageexperiments}
We train two kinds of SSL approaches for visual representations as discussed in Section \ref{sec:SSL_images}. We use ImageNet \cite{deng2009imagenet} dataset with $1.28$M images for training SSL tasks. For the evaluation of these SSL tasks and Multi-SSL models, we investigate on three downstream tasks. 
We use Pascal VOC \cite{everingham2010pascal} on image classification  and MS COCO datasets \cite{lin2014microsoft} on object detection and instance segmentation.

\noindent
\textbf{Settings} 
For image representation learning on ImageNet \cite{deng2009imagenet}, we follow the settings from \cite{chen2020improved,he2020momentum}. A ResNet50 \cite{he2016deep} is adopted as the backbone. The global projection head in MoCov2 \cite{he2020momentum} and dense projection head in DenseCL \cite{wang2021dense} have a output of 512D feature vector and dense 512D feature vectors respectively. 
We adopt SGD as the optimizer and set its weight decay and momentum to $0.0001$ and $0.9$. 
We train for $200$ epochs.
More details are in the supplementary material. 
Let us now move to the standard image downstream tasks.

\noindent
\textbf{Image classification}. We investigate results of image classification on Pascal VOC dataset \cite{everingham2010pascal}. We follow \cite{goyal2019scaling} and train linear SVMs using the feature representations extracted from the frozen encoder pretrained on SSL tasks. Finally, we evaluate on the VOC07 top-1 accuracy as in Table \ref{tab:visual:multiple_SSL}.

\noindent
\textbf{Detection and instance segmentation} We evaluate our pretrained SSL and Multi-SSL models on object detection and instance segmentation. We train a Mask R-CNN detector \cite{he2017mask} using pretrained FPN-backbone on COCO \texttt{train2017} split and evaluate on COCO  \texttt{val2017} split. Table \ref{tab:visual:multiple_SSL} reports the results of object detection and instance segmentation results on COCO dataset.  

\subsection{Learning with single SSL}
\label{sec:sound:single_SSL}

We tabulate the downstream task results of single SSL tasks for sound and image representation learning in Table \ref{tab:sound:multiple_SSL}(a) and \ref{tab:visual:multiple_SSL} respectively. 

\smallskip
\noindent
\textbf{Good performers in Single SSL tasks.}
In Table \ref{tab:sound:multiple_SSL}(a), 
we see that SSL task $\mathcal{A}$ performs best in S$^3$R task compared to SSL tasks $\mathcal{B}$ and $\mathcal{C}$. For instance, $\mathcal{A}$ gets S$^3$R error of $0.2105$ which is $0.039$ and $0.092$ lower than other single SSL counterparts. This can be because aligning binaural sounds with spatial video cues in $\mathcal{A}$, help in learning representations variant to spatial directions. These representations further help in learning the directional resolution of sounds. Coming to other tasks, $\mathcal{B}$ outperforms other two tasks ($\mathcal{A}$ and $\mathcal{C}$) in semantic prediction and video retrieval performance as in Table \ref{tab:sound:multiple_SSL}(a). This is because, learning to align the foreground objects and binaural sounds may have allowed the sound representations to capture the semantics and spatial cues of objects. Coming to visual part, we see that DenseCL performs good in object detection and instance segmentation performance with COCO, compared to MoCov2 as in Table \ref{tab:visual:multiple_SSL} while the latter performs better in image classification on VOC07. This has clear explanation in previous work \cite{wang2021dense} that MoCov2 captures global image features while DenseCL captures the local ones.

\smallskip
\noindent
\textbf{Supervised models vs single SSL.} As we note in Table \ref{tab:sound:multiple_SSL}(a), supervised pretrained models pretrained on AudioSet \cite{hershey2017cnn,gemmeke2017audio}, perform better than single SSL tasks. For \textit{eg.,} $\mathcal{B}$ gets semantic prediction mean IoU of $24.33$ while $\mathcal{A}$ has an S$^3$R error of $0.2105$ and these are less than the supervised counterparts with $26.82$ and $0.2085$ respectively. Contrasting to this, visual SSL of contrastive learning approaches MoCov2 and DenseCL in Table \ref{tab:visual:multiple_SSL}, surpass supervised models that are pretrained on ImageNet \cite{deng2009imagenet}, by $0.2\%$ in VOC07 classification and $1.34$ on $AP$ and $0.82$ on $AP^{mk}$.

\subsection{Learning with Multi-SSL}
\label{sec:sound:multiple_SSL}

Different approaches of Multi-SSL models are tabulated for sound and image representations in Table \ref{tab:sound:multiple_SSL}(b) and Table \ref{tab:visual:multiple_SSL} respectively on their respective downstream tasks. 

\smallskip
\noindent
\textbf{Different Multi-SSL methods.} \textit{Concatenation} and \textit{Multi-task} approach with SSL tasks $\mathcal{B}$ and $\mathcal{C}$, as discussed in Section \ref{sec:SSL_binaural}, achieve $26.37$ and $27.28$ respectively. We further adopt the work of ProgressiveNet \cite{rusu2016progressive} where we use baseline2 approach of \cite{rusu2016progressive} which shows promising performance of $30.38$ on tasks $\mathcal{B}$ \& $\mathcal{C}$ as in Table \ref{tab:sound:multiple_SSL}. 
Then, we introduce incremental learning (IL) that achieves $32.78$ mean IoU with $\mathcal{B}$+$\mathcal{C}$. Further, we see that IL with $\mathcal{B}$+ $\mathcal{C}$+$\mathcal{A}$ scores mean IoU of $34.05$ that outperforms all other methods with a huge margin. For video retrieval and S$^3$R tasks too, we see that IL approach performs better than other Multi-SSL models.
Coming to image downstream tasks, IL approach for image representations performs better than \textit{Multi-task} and \textit{Concatenation}, with an improvement of $0.81$ AP and $1.15$ AP respectively on COCO detection in Table \ref{tab:visual:multiple_SSL}. We notice similar trend with VOC07 accuracy and COCO instance segmentation.

\smallskip
\noindent
\textbf{Multi-SSL improves over single SSL task.}
We see that IL of $\mathcal{B}$+$\mathcal{C}$+$\mathcal{A}$ clearly outperforms best performing single SSL tasks with \textbf{+9.72} mean IoU and \textbf{+2.97\%} video retrieval top-1 accuracy higher than $\mathcal{B}$  and \textbf{0.01} S$^3$R error lower than $\mathcal{A}$ as in Table \ref{tab:sound:multiple_SSL}. Further, we plot single SSL and Multi-SSL IL approach as a tradeoff between different downstream task performance in Figure \ref{fig:tradeoff_seman_VR}. We notice that $\mathcal{B}$+$\mathcal{C}$+$\mathcal{A}$ outbeats single SSL tasks and two task combination in the tradeoff between all downstream tasks.
Further we notice in Table \ref{tab:visual:multiple_SSL} that IL approach has significant advantages over MoCov2 and other recent methods \textit{e.g.} DenseCL \cite{wang2021dense} and DetCo \cite{xie2021detco} by \textbf{+2.83}, \textbf{+1.56} and \textbf{+1.61} $AP$ on COCO detection. 
On instance segmentation, IL outperforms MoCov2 and DenseCL by +1.57 and +0.99 on $AP^{mk}$. On image classification, IL is also $2.06\%$ and $3.27\%$ higher than MoCov2 and DenseCL on VOC07 accuracy. In Figure \ref{fig:teaser}, we see that IL achieves the best performance trade-off on both classification and detection unlike SSL tasks.

\smallskip
\noindent
\textbf{Multi-SSL outperforms supervised counterparts.}  
We notice that Multi-SSL models improve over supervised models both in sound and image downstream tasks. In Table \ref{tab:sound:multiple_SSL}(a,b), Multi-SSL IL method significantly outperforms over supervised models, especially +$7.23$ on mean IoU and -$0.01$ on S$^3$R error. Coming to Table \ref{tab:visual:multiple_SSL}, we see that IL(M+D) is higher than ImageNet supervised models by $2.26\%$ in VOC07 accuracy, 2.57 on $AP_{50}$ and 1.81 on $AP^{mk}$.

\begin{table}[!tb]
\caption{Evaluation of image classification on VOC07, Object detection and instance segmentation on COCO dataset using pretrained Multi-SSL models from ImageNet for 200 epochs, having ResNet50 as the trunk. For downstream, we use Mask R-CNN detector (FPN-backbone).}\label{tab:visual:multiple_SSL}%
\resizebox{1.03\linewidth}{!}{
  \begin{tabular}{ccccccccccccccc}
\toprule 
 \multicolumn{1}{c}{Downstream $\longrightarrow$} & \multicolumn{1}{c}{VOC07} & \multicolumn{3}{c}{COCO detection} &  \multicolumn{3}{c}{COCO instance segm} \\
  Pretrain Tasks& Acc & AP & $AP_{50}$ & $AP_{75}$ & $AP^{mk}$ & $AP_{50}^{mk}$ & $AP_{75}^{mk}$ \\  
\midrule
Supervised & 84.12 & 38.92 & 59.55 & 42.83  & 35.40 & 56.60 & 38.14 \\  
 DetCo \cite{xie2021detco} & 85.19 & 40.21 & 61.11 & 43.84 & 36.36 & 58.12 & 38.97 \\
\midrule
 MoCov2(M)  \cite{he2020momentum} & 84.32 & 38.99  & 59.78 & 42.57 & 35.64 & 56.62 & 38.02 \\
 DenseCL(D) \cite{wang2021dense} & 83.11 & 40.26 & 59.92 & 44.35  &  36.22 & 57.61 & 38.78 \\
\midrule
 MTL(M+D) \cite{wang2021dense} & 85.82 & 41.01 & 60.96 & 44.66 & 36.41 & 57.84 & 39.15 \\ 
 Concat(M+D) & 85.54 & 40.67 & 60.59 & 43.98 & 36.61 & 58.18 & 39.20\\ 
 IL(M+D) & \textbf{86.38} & \textbf{41.82} & \textbf{62.02} & \textbf{45.01} & \textbf{37.21} & \textbf{59.10} & \textbf{39.93}  \\ 
\toprule 
\end{tabular} 
}
\end{table}

\subsection{Ablation studies}
\label{sec:sound:ablation}
\noindent
\textbf{More SSL tasks in Multi-SSL better the performance:}
We perform ablation studies on different combination of SSL tasks in Multi-SSL for sound representations in Table \ref{tab:sound:ablation:multissl}. We showcase the performance of different Multi-SSL methods, \textit{i.e., Multi-task, Concatenation} and IL using 3 downstream tasks.
In Table \ref{tab:sound:ablation:multissl}, Concatenation of $\mathcal{A}$+$\mathcal{B}$ get a mean IoU of $29.55$ and it boosts to $30.14$ when $\mathcal{C}$ is added. This can be observed with $\mathcal{B}$+$\mathcal{C}$ and $\mathcal{A}$+$\mathcal{C}$. 
With \textit{Multi-task} approach, 
we see that joint training of all $3$ SSL tasks achieves $31.21$ mean IoU bettering all other combinations of tasks. Coming to IL approach, $\mathcal{B}$+$\mathcal{C}$ scores $32.76$ mean IoU which is higher than the single task $\mathcal{B}$ and $\mathcal{C}$. Finally, when $\mathcal{A}$ is added to IL model, performance improves to $34.05$ in mean IoU which outbeats 
two tasks combination and single task performance as in Table \ref{tab:sound:ablation:multissl}. We see the same trend with video retrieval top-1 accuracy and S$^3$R error measures as well. This shows that adding more SSL tasks to Multi-SSL approaches improve the downstream performance. 

\smallskip
\noindent
\textbf{Incremental learning approach for general features:} From Figure \ref{fig:teaser} and \ref{fig:tradeoff_seman_VR}, 
we note that single SSL task display good performance with one or two downstream tasks for which they are designed but not on all the tasks. For instance, in Table \ref{tab:sound:multiple_SSL}(a), Task $\mathcal{A}$ performs best in S$^3$R task but poor in other tasks.
Task $\mathcal{B}$ performs good in video retrieval and semantic prediction while $\mathcal{C}$ shows mediocre performance in all three downstream tasks. In contrast, we see in Table \ref{tab:sound:ablation:multissl} that Multi-SSL performs well in all three downstream tasks. 
Figure \ref{fig:tradeoff_seman_VR} more clearly distinguishes the performances of single task and Multi-SSL (IL), displaying the tradeoff between downstream task performance. We note that Multi-SSL (IL) outperforms on both the tasks in all subfigures of Figure \ref{fig:tradeoff_seman_VR}. We see the same trend in Figure \ref{fig:teaser} showing its generalizability. In addition, Multi-SSL IL shows advantages over \textit{Multi-task} and \textit{Concatenation} here.
Overall, above experiments indicate that Multi-SSL approach extract more generic features that perform well in all the downstream tasks.


\subsection{Limitations and future works}
Our paper investigates a handful of approaches of Multi-SSL.
We strongly believe that this work will open up more directions to combine SSL tasks and this can raise SSL benchmarks.
Further, in our work, the tasks are learned sequentially in Multi-SSL models and this results in more training time than single SSL tasks. 
For Multi-SSL models, we limit to $3$ SSL tasks during the experiments.
In future, it would be interesting to bring in more tasks and examine affinity mapping between combination of SSL tasks and downstream tasks.

\section{Conclusion}

This work proposes different approaches of 
Multi-SSL that learn to combine multiple SSL tasks. Experiments on OmniAudio dataset show that Multi-SSL via incremental learning outperforms all single SSL tasks and supervised models on the downstream tasks of semantic prediction, video retrieval, and spatial sound super resolution. We see similar trends of Multi-SSL on image representation learning using MoCov2 and DenseCL as SSL tasks. It demonstrates state-of-the-art performance on VOC classification, COCO detection and instance segmentation.

\vspace{2mm}
\noindent
\textbf{Acknowledgement}:
This work is funded by Toyota Motor Europe via the research project TRACE-Zurich.

{\small
\bibliographystyle{ieee}
\bibliography{egbib}
}

\end{document}